\documentclass[a4paper,fleqn]{cas-dc}
\usepackage[numbers,sort&compress]{natbib}
\usepackage{amsmath,amssymb,amsfonts}
\usepackage{graphicx}
\usepackage{booktabs}
\usepackage{multirow}
\usepackage{xcolor}
\usepackage{url}
\newcommand{\method}{ISTS}

\shorttitle{Interpretable structural temperature scaling}
\shortauthors{C. Huang}

\ExplSyntaxOn

\cs_set:Npn \__first_footerline:
  { \group_begin: \small \sffamily \__short_authors: :~
    { \rmfamily \itshape Preprint~ under~ consideration~ at~ Pattern~ Recognition~ Letters } \group_end: }
\ExplSyntaxOff
\begin{document}
\let\WriteBookmarks\relax
\def\floatpagepagefraction{1}
\def\textpagefraction{.001}

\title [mode = title]{Interpretable structural temperature scaling for calibrating graph neural networks and point-cloud segmentation}

\author[1]{Chenyi Huang}[type=editor]
\cormark[1]
\ead{chenyi.huang.25@ucl.ac.uk}
\ead{hancyih@gmail.com}
\affiliation[1]{organization={Department of Computer Science, University College London},
                addressline={Gower Street},
                city={London},
                postcode={WC1E 6BT},
                country={United Kingdom}}
\cortext[cor1]{Corresponding author}

\begin{abstract}
Graph-aware calibrators make graph neural networks (GNNs) more trustworthy by predicting a temperature for each node, usually with auxiliary networks whose decisions are hard to inspect. We propose Interpretable Structural Temperature Scaling (ISTS), a twelve-parameter linear model that sets each node's temperature from named descriptors of its confidence, neighbourhood and closeness to labelled nodes, so every temperature traces back to descriptor contributions. ISTS preserves predictions, and cross-validated regularisation lets the calibration data decide how far the model departs from temperature scaling (TS). Its explanations are faithful at the level of node groups: on test nodes, ISTS softens groups that a global temperature leaves over-confident and sharpens under-confident ones (Spearman correlation 0.67 over 246 groups). On seven graphs with two backbones and eleven baselines, ISTS has the best average rank in negative log-likelihood (NLL) and Brier score and lowers NLL relative to TS by 1.8\%, more than twice the gain of an identically fitted entropy-based temperature. It ranked first again on four held-out graphs under a protocol fixed in advance. From about 100 calibration nodes on, ISTS is a cheap and inspectable choice, and on point-cloud segmentation it significantly improves NLL, Brier score and the ranking of points for rejection.
\end{abstract}


\maketitle

\section{Introduction}
\label{sec:intro}

Graph neural networks (GNNs) such as GCN~\citep{kipf2017gcn} and GAT~\citep{velickovic2018gat}
are the standard tool for semi-supervised node classification. When their outputs feed
decisions, the predicted probabilities must be trustworthy: among nodes predicted with
confidence $c$, a fraction $c$ should be correct~\citep{guo2017calibration}. GNNs are often
miscalibrated and, unlike modern image classifiers, tend to be
\emph{under}-confident~\citep{wang2021cagcn,hsu2022gats}.

Post-hoc calibration fixes this without retraining. It fits a small map from the frozen logits
to probabilities on a held-out labelled calibration set. Temperature scaling
(TS)~\citep{guo2017calibration} uses one scalar for all inputs. On graphs, however, the right
correction differs between nodes: it depends on a node's distance to the labelled set, on how
its confidence compares with its neighbours', and on whether its neighbourhood agrees with its
prediction~\citep{hsu2022gats,liu2022rbs}. Graph-aware calibrators therefore predict node-wise
temperatures with an auxiliary GCN~\citep{wang2021cagcn}, an attention
layer~\citep{hsu2022gats}, a gated mixture of GNN experts~\citep{zhuang2025gets}, or graph
wavelets~\citep{li2026wats}. These models are trained on a calibration set that is small by
construction, and they offer little insight into \emph{why} a node's confidence was changed.

The main factors behind GNN miscalibration are already known~\citep{hsu2022gats,liu2022rbs}, and we
show that combining them linearly yields a competitive calibrator whose decisions can be read off. We
\mbox{propose} \textbf{Interpretable Structural Temperature Scaling (\method{})}, which maps eleven
standardised, human-readable node descriptors to a node-wise temperature through a generalised
linear model with twelve parameters. Our contributions are:
\begin{enumerate}
\item An accuracy-preserving post-hoc calibrator with inspectable decisions, whose size does not depend on
the number of classes or the graph. It turns known miscalibration
factors into explicit regressors, and its regularisation, selected with a one-standard-error rule,
lets the calibration data decide how much node-wise structure to use.
\item A controlled comparison with eleven baselines, including the closely related HoTS and WATS and
entropy- and logit-based temperatures (HTS, LTS) fitted with the same pipeline, on seven graphs and
two backbones, and a confirmation on four held-out graphs under a frozen protocol. \method{} obtains
the best average rank on NLL and Brier score in both, and is about $90\times$ faster than GATS on the largest graph.
\item An analysis of when and why structure helps: a test showing that on unseen nodes the learned
temperatures move in the direction of the miscalibration a global temperature leaves, a descriptor ablation,
and a calibration-set-size study that locates where node-wise structure starts to pay off (about 100
calibration nodes on the citation graphs).
\item A transfer study to per-point calibration of a point-cloud segmentation network, the kind of
perception model robots rely on. Keeping the temperature model and adapting only the descriptors, \method{} improves NLL, Brier score and the
selective-prediction ranking of points, also under simulated sensor degradation.
\end{enumerate}

\raggedbottom
\section{Related Work}
\label{sec:related}

\subsection{Post-hoc calibration}
In temperature scaling (TS)~\citep{guo2017calibration}, one learned scalar divides all logits.
Vector/matrix scaling and Dirichlet calibration~\citep{kull2019dirichlet} add class-wise
parameters, ensemble temperature scaling (ETS)~\citep{zhang2020mixnmatch} mixes the
temperature-scaled, uncalibrated and uniform distributions, and sample-dependent
temperatures have been studied for i.i.d.\ data, e.g.\ as functions of the predictive entropy
or the logits~\citep{joy2023adaptive,balanya2024ats}.

\subsection{Calibration of GNNs}
GNNs were first reported to be miscalibrated by \citet{teixeira2019miscalibrated}.
\citet{wang2021cagcn} found that semi-supervised GNNs tend to be
\emph{under}-confident and proposed CaGCN, an auxiliary GCN that outputs node-wise
temperatures. Ratio-binned scaling (RBS)~\citep{liu2022rbs} fits one temperature per bin of
an estimated same-class-neighbour ratio. \citet{hsu2022gats} identified five factors
behind GNN miscalibration: general under-confidence, diversity of predictive distributions,
distance to training nodes, relative confidence, and neighbourhood similarity. They built
graph attention temperature scaling (GATS) around these factors. Later post-hoc methods use
prototype similarity and homophily (SimCalib~\citep{tang2024simcalib}), groupings by
neighbourhood similarity and confidence (Simi-Mailbox~\citep{seo2025simimailbox}), a gated
mixture of GNN temperature experts (GETS~\citep{zhuang2025gets}), or graph wavelets
(WATS~\citep{li2026wats}). HoTS~\citep{tae2026hots} lowers the temperature of nodes with high estimated local homophily. Train-time remedies~\citep{wang2022gcl,wang2024auls} are complementary to post-hoc calibration.

\subsection{Positioning of \method{}}
\method{} is closest to HoTS: both keep the predicted class and set a node-wise temperature from
confidence and local structure. HoTS derives its functional form from theory and estimates homophily
with a trained GNN (about $4.6\times10^4$ weights on Cora). \method{} learns a linear form over a dozen
descriptors that operationalise the factors of \citet{hsu2022gats}, plus a
label-propagation~\citep{zhou2004lgc} signal, without an auxiliary network. Its coefficients can be read
directly, and its one-standard-error rule moves it towards TS, its limit as $\lambda\to\infty$, as far as
the calibration data require. RBS is a piecewise-constant counterpart of it.

\section{Method}
\label{sec:method}

\subsection{Problem setting}
Let $\mathcal{G}=(\mathcal{V},\mathcal{E})$ be a graph with labels $y_i\in\{1,\dots,K\}$ known
for a training set $\mathcal{V}_{\mathrm{tr}}$ and a disjoint calibration set
$\mathcal{V}_{\mathrm{cal}}$. A GNN trained on $\mathcal{V}_{\mathrm{tr}}$ outputs logits
$\mathbf{z}_i\in\mathbb{R}^K$, probabilities $\mathbf{p}_i=\mathrm{softmax}(\mathbf{z}_i)$,
prediction $\hat y_i=\arg\max_k p_{ik}$ and confidence $c_i=\max_k p_{ik}$. A post-hoc
calibrator keeps the GNN frozen and maps $\mathbf{z}_i$ (and possibly the graph) to calibrated
probabilities $\hat{\mathbf{p}}_i$, using labels in $\mathcal{V}_{\mathrm{cal}}$ and, as
in \citet{hsu2022gats}, the identity and labels of $\mathcal{V}_{\mathrm{tr}}$. Metrics are defined in Section~S3 of the supplementary material.

\subsection{Node-wise temperature model}
\method{} gives node $i$ the temperature and calibrated probabilities
\begin{equation}
T_i=\operatorname{softplus}\!\bigl(b+\mathbf{w}^{\top}\tilde{\boldsymbol{\phi}}_i\bigr)+\tau_{\min},\qquad
\hat{\mathbf{p}}_i=\mathrm{softmax}(\mathbf{z}_i/T_i),
\label{eq:ists}
\end{equation}
with $\tau_{\min}=0.05$ and $(b,\mathbf{w})\in\mathbb{R}\times\mathbb{R}^{11}$ the only learnable
parameters. The descriptors $\boldsymbol{\phi}_i\in\mathbb{R}^{11}$ are standardised,
\begin{equation}
\tilde\phi_{ij}=(\phi_{ij}-\mu_j)/\sigma_j ,
\label{eq:std}
\end{equation}
with $\mu_j$ and $\sigma_j$ the mean and standard deviation of $\phi_j$ over all nodes.
The model has three useful properties:
\begin{equation}
\begin{gathered}
\arg\max_k\hat p_{ik}=\hat y_i,\qquad
\mathbf{w}=\mathbf{0}\;\Rightarrow\;T_i\equiv T,\\
\operatorname{sign}\bigl(\partial T_i/\partial\tilde\phi_{ij}\bigr)=\operatorname{sign}(w_j).
\end{gathered}
\label{eq:props}
\end{equation}
Accuracy is preserved because $T_i>0$ rescales all logits of a node equally; with $\mathbf{w}=\mathbf{0}$
the model is TS; and a positive $w_j$ softens nodes with large $\tilde\phi_{ij}$, a negative one
sharpens them.

\subsection{Structural descriptors}
Table~\ref{tab:params} defines the 11 descriptors, with $\mathcal{N}(i)$ the neighbours of $i$,
$d_i=|\mathcal{N}(i)|$ and $\delta_i$ the hop distance to the nearest training node. They form four
groups that operationalise the miscalibration factors of \citet{hsu2022gats}: \emph{confidence}
($\phi_1,\phi_2$), \emph{neighbourhood} ($\phi_3$--$\phi_6$; $\phi_3$ is related to the soft-support
ratio of RBS~\citep{liu2022rbs}), \emph{label propagation} ($\phi_7,\phi_8$) and \emph{distance}
($\phi_9$--$\phi_{11}$). The LP scores come from personalised propagation of the training
labels~\citep{zhou2004lgc},
\begin{equation}
\mathbf{Q}^{(l+1)}=\alpha\,\mathbf{D}^{-1/2}\mathbf{A}\mathbf{D}^{-1/2}\mathbf{Q}^{(l)}+(1-\alpha)\mathbf{Y}_{\mathrm{tr}},
\label{eq:lp}
\end{equation}
with $\mathbf{Q}^{(0)}=\mathbf{Y}_{\mathrm{tr}}$ (one-hot training labels), $\alpha=0.9$, $L=10$ and
$\bar{\mathbf q}_i=\mathbf q^{(L)}_i/\|\mathbf q^{(L)}_i\|_1$ (Section~S3). LP sees only the structure and the training labels, so disagreement between
$\bar{\mathbf q}_i$ and the GNN signals likely errors.

\begin{table}[t]
\centering
\caption{The 11+1 parameters of \method{}: the bias $b$ and one weight $w_j$ per descriptor $\phi_j$.
$\delta_i$: hop distance to the nearest training node.}
\label{tab:params}
\footnotesize\setlength{\tabcolsep}{5pt}
\begin{tabular}{@{}lll@{}}
\toprule
Symbol & Descriptor & Definition\\
\midrule
$b$ & bias & TS temperature level when $\mathbf{w}=\mathbf{0}$\\
$\phi_1$ & confidence & $c_i$\\
$\phi_2$ & entropy & $H(\mathbf{p}_i)/\log K$\\
$\phi_3$ & agreement & $\frac{1}{d_i}\sum_{j\in\mathcal{N}(i)}\mathbb{1}[\hat y_j=\hat y_i]$\\
$\phi_4$ & relative confidence & $c_i-\frac{1}{d_i}\sum_{j\in\mathcal{N}(i)}c_j$\\
$\phi_5$ & similarity & $\frac{1}{d_i}\sum_{j\in\mathcal{N}(i)}\cos(\mathbf{p}_i,\mathbf{p}_j)$\\
$\phi_6$ & log-degree & $\log(1+d_i)$\\
$\phi_7$ & LP agreement & $\bar q_{i,\hat y_i}$\\
$\phi_8$ & LP confidence & $\max_k\bar q_{ik}$\\
$\phi_9$--$\phi_{11}$ & distance & $\mathbb{1}[\delta_i=1]$, $\mathbb{1}[\delta_i=2]$, $\mathbb{1}[\delta_i\geq3]$\\
\bottomrule
\end{tabular}
\end{table}

\subsection{Fitting and complexity}
$(b,\mathbf{w})$ minimise the penalised calibration NLL
\begin{equation}
\mathcal{L}_\lambda(b,\mathbf{w})=\frac{1}{|\mathcal{V}_{\mathrm{cal}}|}\sum_{i\in\mathcal{V}_{\mathrm{cal}}}-\log\hat p_{i,y_i}+\lambda\|\mathbf{w}\|_2^2
\label{eq:loss}
\end{equation}
with L-BFGS, warm-started from TS ($\mathbf{w}=\mathbf{0}$). As $b$ is not penalised,
$\lambda\to\infty$ gives TS. With held-out per-node losses $\ell_i(\lambda)$ from 5-fold
cross-validation inside $\mathcal{V}_{\mathrm{cal}}$, $\lambda^{\ast}=\arg\min_\lambda\bar\ell(\lambda)$ and
$\Delta_i(\lambda)=\ell_i(\lambda)-\ell_i(\lambda^{\ast})$, the one-standard-error
rule~\citep{hastie2009elements} selects
\begin{equation}
\hat\lambda=\max\Bigl\{\lambda\in\Lambda:\ \bar\Delta(\lambda)\le \frac{\operatorname{sd}(\Delta(\lambda))}{\sqrt{|\mathcal{V}_{\mathrm{cal}}|}}\Bigr\}
\label{eq:onese}
\end{equation}
over $\Lambda=\{10^{-3},10^{-2},\dots,10^{2}\}$, the strongest shrinkage the calibration data cannot reject; the rule needs no per-dataset tuning. Descriptors cost $O(L|\mathcal{E}|K+nK)$ and each L-BFGS
step $O(|\mathcal{V}_{\mathrm{cal}}|(K+d))$; the 12 parameters do not depend on $K$ or the graph size
(baselines: Table~\ref{tab:rank}).

\section{Experiments}
\label{sec:exp}

\subsection{Setup}
\textbf{Data and backbones.} We use Cora, Citeseer and Pubmed~\citep{sen2008collective,yang2016planetoid},
Amazon Computers and Photo, and Coauthor CS and Physics~\citep{shchur2018pitfalls}, with the
public splits of \citet{hsu2022gats}: in each of five random splits, 15\% of the nodes are labelled
and divided into three folds, one of which is the calibration set (5\% of the nodes, e.g.\ 134 on
Cora), and the other 85\% are test nodes. Two-layer GCN~\citep{kipf2017gcn} and
GAT~\citep{velickovic2018gat} follow the GATS code base in PyTorch Geometric~\citep{fey2019pyg},
giving 15 split--fold runs per setting (details in Section~S4). Only the five splits are
independent, as the folds of a split share its test nodes.

\textbf{Baselines.} TS, VS~\citep{guo2017calibration}, ETS~\citep{zhang2020mixnmatch},
CaGCN~\citep{wang2021cagcn} and GATS~\citep{hsu2022gats} use the official GATS implementation and
its per-dataset configurations. HoTS~\citep{tae2026hots} and GETS~\citep{zhuang2025gets} are our
PyTorch Geometric ports of the official DGL code with the released configurations (Sections~S2
and~S6). RBS~\citep{liu2022rbs} follows its original definition. HTS and LTS~\citep{balanya2024ats}
map the entropy (2 parameters) or the logits ($K{+}1$) to a temperature and are fitted with the
\method{} pipeline, so they differ from it only in their inputs. WATS~\citep{li2026wats} uses the paper's
settings and trains to convergence on the calibration loss, the stronger of the two training rules we compared (Section~S12).

\textbf{Metrics.} ECE~\citep{naeini2015bbq} (15 equal-width bins), NLL and Brier score on the test nodes; as binned
ECE is noisy and biased~\citep{kumar2019verified}, the proper scoring rules get equal weight.
Only VS and GETS can change predictions.

\textbf{Design history.} A pilot version of \method{} selected $\lambda$ by plain cross-validation over
$\{10^{-3},\dots,1\}$. To handle very small calibration sets (Sec.~\ref{sec:small}), we added $10$, $10^{2}$
and the one-standard-error rule after seeing the pilot test results (``pilot'' in Table~S3); Section~\ref{sec:confirm}
tests the final design on unseen graphs.

\begin{table*}[t]
\centering
\caption{Test NLL ($\times100$; mean$\pm$std over 15 runs) after post-hoc calibration. ECE and Brier score are in Tables~S1 and~S2 of the supplementary material. Lower is better; best calibrated result in bold and underlined (Uncal.\ excluded). RBS follows the original definition (equal-width bins); HTS and LTS are fitted with the same pipeline as \method{}. VS and GETS can change predictions; all other calibrators preserve accuracy.}
\label{tab:main}
\setlength{\tabcolsep}{1.9pt}\footnotesize
\begin{tabular}{llccccccccccccc}
\toprule
Dataset & Model & Uncal. & TS & VS & ETS & HTS & LTS & RBS & CaGCN & GATS & GETS & HoTS & WATS & \method{}  \\
\midrule
Cora & GCN & 62.9\tiny{$\pm$5.2} & 56.4\tiny{$\pm$3.1} & 57.6\tiny{$\pm$4.6} & 55.9\tiny{$\pm$2.9} & 55.8\tiny{$\pm$3.1} & 56.0\tiny{$\pm$2.9} & 57.6\tiny{$\pm$3.8} & 65.4\tiny{$\pm$6.7} & 55.8\tiny{$\pm$3.2} & 65.0\tiny{$\pm$6.4} & \textbf{\underline{54.3\tiny{$\pm$3.2}}} & 56.1\tiny{$\pm$2.9} & 55.8\tiny{$\pm$2.7} \\
Citeseer & GCN & 91.8\tiny{$\pm$8.2} & 87.6\tiny{$\pm$3.0} & 87.3\tiny{$\pm$2.4} & 87.0\tiny{$\pm$2.8} & 87.2\tiny{$\pm$2.8} & 87.2\tiny{$\pm$2.8} & 90.0\tiny{$\pm$7.4} & 94.5\tiny{$\pm$5.7} & \textbf{\underline{86.7\tiny{$\pm$2.9}}} & 91.6\tiny{$\pm$4.6} & 87.2\tiny{$\pm$3.3} & 87.4\tiny{$\pm$2.7} & 87.1\tiny{$\pm$2.8} \\
Pubmed & GCN & 39.5\tiny{$\pm$1.2} & 36.7\tiny{$\pm$0.6} & 36.8\tiny{$\pm$0.6} & 36.5\tiny{$\pm$0.7} & 36.5\tiny{$\pm$0.7} & 36.4\tiny{$\pm$0.8} & 37.0\tiny{$\pm$1.0} & \textbf{\underline{35.8\tiny{$\pm$0.6}}} & 36.4\tiny{$\pm$0.6} & 36.9\tiny{$\pm$0.6} & 36.5\tiny{$\pm$0.6} & 36.7\tiny{$\pm$0.6} & 35.9\tiny{$\pm$0.7} \\
Computers & GCN & 42.8\tiny{$\pm$1.4} & 42.8\tiny{$\pm$1.4} & 42.8\tiny{$\pm$1.4} & 40.5\tiny{$\pm$1.4} & 41.4\tiny{$\pm$1.9} & 42.3\tiny{$\pm$2.0} & 40.4\tiny{$\pm$1.4} & 43.2\tiny{$\pm$3.2} & 42.2\tiny{$\pm$1.7} & 51.7\tiny{$\pm$7.3} & 41.9\tiny{$\pm$1.3} & \textbf{\underline{37.6\tiny{$\pm$1.4}}} & 38.5\tiny{$\pm$1.4} \\
Photo & GCN & 28.6\tiny{$\pm$1.0} & 28.8\tiny{$\pm$1.1} & 28.9\tiny{$\pm$0.8} & 27.0\tiny{$\pm$1.1} & 28.6\tiny{$\pm$1.2} & 28.7\tiny{$\pm$1.3} & 27.6\tiny{$\pm$1.3} & 40.4\tiny{$\pm$10.0} & 28.5\tiny{$\pm$1.1} & 44.4\tiny{$\pm$13.9} & 28.4\tiny{$\pm$1.1} & \textbf{\underline{26.4\tiny{$\pm$2.5}}} & 27.8\tiny{$\pm$2.3} \\
CS & GCN & 22.1\tiny{$\pm$1.1} & 21.4\tiny{$\pm$0.4} & 21.7\tiny{$\pm$0.4} & 21.4\tiny{$\pm$0.4} & 21.4\tiny{$\pm$0.4} & 21.4\tiny{$\pm$0.4} & 21.5\tiny{$\pm$0.4} & 26.4\tiny{$\pm$6.0} & 21.4\tiny{$\pm$0.4} & 22.7\tiny{$\pm$0.8} & \textbf{\underline{21.2\tiny{$\pm$0.4}}} & 21.4\tiny{$\pm$0.4} & 21.4\tiny{$\pm$0.4} \\
Physics & GCN & 12.0\tiny{$\pm$0.5} & 11.9\tiny{$\pm$0.4} & 12.0\tiny{$\pm$0.4} & 11.9\tiny{$\pm$0.4} & 11.9\tiny{$\pm$0.4} & 11.9\tiny{$\pm$0.4} & 12.2\tiny{$\pm$0.5} & 13.3\tiny{$\pm$1.2} & 11.9\tiny{$\pm$0.4} & 12.2\tiny{$\pm$0.4} & 12.0\tiny{$\pm$0.4} & 11.9\tiny{$\pm$0.4} & \textbf{\underline{11.9\tiny{$\pm$0.4}}} \\
\midrule
Cora & GAT & 59.2\tiny{$\pm$2.7} & 52.0\tiny{$\pm$2.7} & 52.8\tiny{$\pm$3.8} & 51.7\tiny{$\pm$2.4} & 51.5\tiny{$\pm$2.6} & 51.8\tiny{$\pm$2.5} & 53.4\tiny{$\pm$3.3} & 58.1\tiny{$\pm$4.6} & 52.0\tiny{$\pm$2.5} & 56.5\tiny{$\pm$5.9} & \textbf{\underline{50.9\tiny{$\pm$2.8}}} & 51.8\tiny{$\pm$2.5} & 51.8\tiny{$\pm$2.4} \\
Citeseer & GAT & 101.8\tiny{$\pm$17.3} & 89.0\tiny{$\pm$6.1} & \textbf{\underline{87.2\tiny{$\pm$2.4}}} & 87.9\tiny{$\pm$4.4} & 88.3\tiny{$\pm$5.4} & 88.7\tiny{$\pm$6.8} & 89.6\tiny{$\pm$7.0} & 92.3\tiny{$\pm$5.8} & 88.3\tiny{$\pm$5.8} & 89.8\tiny{$\pm$3.1} & 88.7\tiny{$\pm$6.2} & 88.5\tiny{$\pm$5.3} & 88.3\tiny{$\pm$5.4} \\
Pubmed & GAT & 44.0\tiny{$\pm$1.5} & 38.8\tiny{$\pm$0.6} & 38.7\tiny{$\pm$0.6} & 38.7\tiny{$\pm$0.6} & 38.7\tiny{$\pm$0.6} & 38.7\tiny{$\pm$0.6} & 38.9\tiny{$\pm$0.7} & \textbf{\underline{38.4\tiny{$\pm$0.6}}} & 38.7\tiny{$\pm$0.6} & 38.6\tiny{$\pm$0.5} & 38.8\tiny{$\pm$0.6} & 38.7\tiny{$\pm$0.6} & 38.5\tiny{$\pm$0.5} \\
Computers & GAT & 38.1\tiny{$\pm$2.9} & 37.9\tiny{$\pm$2.5} & 37.8\tiny{$\pm$2.4} & 37.2\tiny{$\pm$2.5} & 37.7\tiny{$\pm$2.6} & 37.9\tiny{$\pm$2.7} & 37.1\tiny{$\pm$2.5} & 40.0\tiny{$\pm$3.2} & 37.8\tiny{$\pm$2.5} & 46.8\tiny{$\pm$6.9} & 37.6\tiny{$\pm$2.5} & \textbf{\underline{36.8\tiny{$\pm$2.5}}} & 36.9\tiny{$\pm$2.8} \\
Photo & GAT & 26.7\tiny{$\pm$1.1} & 26.8\tiny{$\pm$1.1} & 26.6\tiny{$\pm$1.1} & 26.2\tiny{$\pm$1.4} & 26.7\tiny{$\pm$1.3} & 26.7\tiny{$\pm$1.3} & 31.0\tiny{$\pm$18.3} & 32.7\tiny{$\pm$4.9} & 26.7\tiny{$\pm$1.3} & 47.2\tiny{$\pm$27.6} & 26.8\tiny{$\pm$1.2} & 26.7\tiny{$\pm$2.1} & \textbf{\underline{26.2\tiny{$\pm$1.4}}} \\
CS & GAT & 24.3\tiny{$\pm$0.5} & 24.3\tiny{$\pm$0.5} & 24.2\tiny{$\pm$0.4} & 24.2\tiny{$\pm$0.4} & 24.2\tiny{$\pm$0.4} & 24.3\tiny{$\pm$0.4} & 24.2\tiny{$\pm$0.5} & 27.6\tiny{$\pm$4.5} & 24.2\tiny{$\pm$0.5} & 26.0\tiny{$\pm$1.4} & \textbf{\underline{24.0\tiny{$\pm$0.5}}} & 24.2\tiny{$\pm$0.5} & 24.1\tiny{$\pm$0.4} \\
Physics & GAT & 12.9\tiny{$\pm$0.4} & 12.9\tiny{$\pm$0.4} & \textbf{\underline{12.8\tiny{$\pm$0.3}}} & 12.9\tiny{$\pm$0.4} & 12.9\tiny{$\pm$0.4} & 12.8\tiny{$\pm$0.4} & 13.1\tiny{$\pm$0.6} & 13.7\tiny{$\pm$0.7} & 12.8\tiny{$\pm$0.4} & 13.1\tiny{$\pm$0.4} & 12.9\tiny{$\pm$0.4} & 12.8\tiny{$\pm$0.4} & 12.8\tiny{$\pm$0.4} \\
\bottomrule
\end{tabular}
\end{table*}

\subsection{Main Results}
Table~\ref{tab:main} gives NLL (ECE and Brier score: Tables~S1--S2), Table~\ref{tab:rank} average
ranks, run time and parameters. A setting is a consistent win (W) or loss (L) for \method{} only if
the fold-averaged differences of all five splits agree, mixed (M) otherwise. With
five independent splits, a two-sided exact sign test cannot attain $p<0.05$ (minimum
$p=0.0625$); we therefore report W/M/L descriptively (Tables~S5--S6).

\textbf{Calibration quality.} \method{} has the best average rank for NLL (2.86) and Brier score
(3.36), and a baseline is consistently better on NLL in only four of 154 comparisons. WATS, the
strongest graph-aware baseline here, reaches a similar mean NLL gain over TS (1.86\%; \method{}: 1.80\%),
concentrated in GCN on Computers and Photo (NLL $\times100$ on Computers: 37.6, \method{} 38.5, TS 42.8);
the gains of \method{} are spread more evenly, giving it the better rank (2.86 vs.\ 4.79). On ECE,
\method{} ranks second (3.64), after GATS (3.21).

\textbf{Value of the structural descriptors.} The structural descriptors carry most of the gain of
\method{}. HTS and LTS set a node-wise temperature from the node's own entropy or logits; fitted with the
same pipeline and the same frozen logits, they differ from \method{} only in these inputs. They reduce NLL
relative to TS by 0.68\% and 0.36\%. \method{} gains 1.80\%, beats HTS in all five splits in 7 of 14
settings on NLL and Brier score (mainly on Computers, Photo and Pubmed) and is never consistently beaten
by it. The ablation (Table~\ref{tab:abl}) agrees: confidence descriptors alone give 0.63\%, and every
structural group adds to this; the groups overlap, so removing any single one costs 0.07--0.24 points.
Compared with HoTS, the closest method, \method{} is consistently better on NLL on Computers, Pubmed
(both backbones), Photo (GAT) and Physics (GCN), and HoTS on Cora and CS (GCN).

\textbf{Computational cost.} \method{} takes 0.3--0.8\,s per run on one CPU thread, including the
30 fits that select $\lambda$; on Physics it is about $90\times$ faster than GATS (Table~\ref{tab:rank}).

\begin{table}[t]
\centering
\caption{Average rank over the 14 settings (lower is better), mean fitting time in seconds on one CPU thread (Cora / Physics, averaged over backbones) and number of fitted parameters on Cora. Split-level W/M/L counts against every baseline are in Table~S5 of the supplementary material.}
\label{tab:rank}
\footnotesize\setlength{\tabcolsep}{3pt}
\begin{tabular}{lcccrr}
\toprule
 & \multicolumn{3}{c}{avg.\ rank} & time (s) & params \\
Method & ECE & NLL & Brier & Cora/Ph. & Cora \\
\midrule
TS & 8.14 & 8.64 & 7.29 & 0.1/0.2 & 1 \\
VS & 8.29 & 6.86 & 5.86 & 0.7/0.3 & 14 \\
ETS & 4.71 & 4.71 & 6.93 & 0.1/0.2 & 4 \\
HTS & 4.64 & 5.07 & 6.00 & 0.2/0.4 & 2 \\
LTS & 6.57 & 5.86 & 5.43 & 0.3/0.5 & 8 \\
RBS & 7.71 & 8.50 & 7.43 & 0.2/0.3 & 2--4 \\
CaGCN & 9.50 & 10.14 & 9.71 & 1.5/44.8 & 145 \\
GATS & 3.21 & 4.64 & 4.29 & 2.1/60.9 & 60 \\
GETS & 10.14 & 10.57 & 10.50 & 21.2/262.6 & 140k \\
HoTS & 6.64 & 5.36 & 6.86 & 1.5/17.4 & 46k \\
WATS & 4.79 & 4.79 & 4.36 & 0.2/6.7 & 113 \\
\method{} & 3.64 & 2.86 & 3.36 & 0.4/0.7 & 12 \\
\bottomrule
\end{tabular}
\end{table}

\begin{table}[t]
\centering
\caption{Descriptor ablation, averaged over the 14 settings: test ECE and relative change in test NLL with respect to TS (negative is better). All variants use the same fitting pipeline. Further variants and HTS/LTS: Table~S3 of the supplementary material.}
\label{tab:abl}
\footnotesize
\begin{tabular}{lcc}
\toprule
Variant & ECE (\%) & $\Delta$NLL vs.\ TS (\%) \\
\midrule
full (11 descriptors) & 1.89 & $-1.80$ \\
confidence only & 2.02 & $-0.63$ \\
w/o neighbourhood & 1.89 & $-1.56$ \\
w/o LP group & 1.90 & $-1.69$ \\
w/o distance & 1.94 & $-1.73$ \\
TS (no descriptors) & 2.14 & $+0.00$ \\
\bottomrule
\end{tabular}
\end{table}

\subsection{Independent Confirmation}
\label{sec:confirm}
As the final design saw pilot test results, we froze the code and recorded a protocol (data, new
stratified splits of the same design, methods, analysis, one prediction; Section~S13) before any run
on unseen graphs: CoraFull, DBLP~\citep{bojchevski2018g2g} and WikiCS~\citep{mernyei2020wikics}, and
heterophilous Amazon-ratings~\citep{platonov2023critical} as a stress test. The prediction, lower NLL
than TS in at least four of the six primary settings, held in all six, and in each of them all five
splits favour \method{} over TS and HTS on NLL and Brier score. \method{} ranks first on both
(Table~\ref{tab:confirm}); on ECE its mean rank is 8.17 of 13. The largest gains occur on WikiCS, whose
backbones train poorly under the unchanged pipeline (accuracy 57\%/29\% for GCN/GAT); without WikiCS
(a post-hoc check) \method{} still ranks first on NLL, with a 2.2\% gain over TS. On Amazon-ratings all
calibrators stay within 0.7\% of TS in NLL (\method{}: 0.4\%).

\begin{table}[t]
\centering
\caption{Independent confirmation on CoraFull, DBLP and WikiCS (six settings, 15 runs each), run under a protocol fixed before any result was seen: average rank among the 13 calibrators and mean relative NLL change with respect to TS. All methods: Table~S17.}
\label{tab:confirm}
\footnotesize\setlength{\tabcolsep}{4pt}
\begin{tabular}{lcccc}
\toprule
 & \multicolumn{3}{c}{avg.\ rank} & $\Delta$NLL \\
Method & NLL & Brier & ECE & vs.\ TS (\%) \\
\midrule
\method{} & 1.17 & 1.50 & 8.17 & $-5.23$ \\
VS & 5.00 & 3.67 & 9.00 & $-2.43$ \\
LTS & 5.67 & 4.67 & 5.50 & $-2.12$ \\
HTS & 6.00 & 6.00 & 4.33 & $-1.54$ \\
HoTS & 6.67 & 7.17 & 7.17 & $-0.68$ \\
WATS & 7.00 & 7.67 & 5.50 & $-0.46$ \\
GATS & 8.50 & 8.67 & 7.83 & $-0.01$ \\
TS & 10.17 & 10.17 & 7.00 & $+0.00$ \\
\bottomrule
\end{tabular}
\end{table}

\begin{figure}[pos=t]
\centering
\includegraphics[width=\columnwidth]{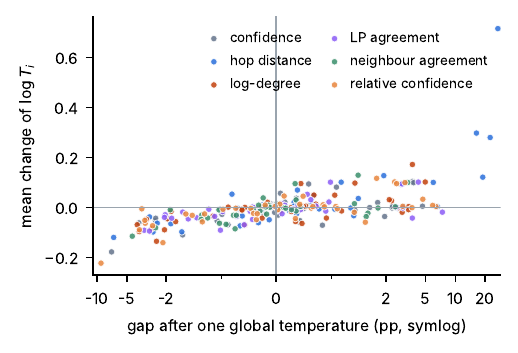}
\caption{Faithfulness of \method{}: per setting and group of test nodes (grouped by one descriptor), the gap
left by one global temperature (confidence minus accuracy) against \method{}'s mean change of log temperature.
Points in the upper-right (over-confident, softened) and lower-left (under-confident, sharpened) quadrants are
corrected in the right direction.}
\label{fig:faith}
\end{figure}

\subsection{What the Model Learns}
\label{sec:coef}
\emph{When structure is used.} The fitted $\lambda$ shows where the calibration data support node-wise
structure. On Computers, where \method{} gains most, the one-standard-error rule selects $\lambda<100$ and thus
node-wise temperatures in 26 of 30 runs; on Citeseer, where no node-wise calibrator gains much over TS, it picks $\lambda=100$,
which makes the temperatures nearly constant, in 28 of 30 (114 of 210 runs overall). The amount of
structure thus follows its usefulness, given enough calibration data (Section~\ref{sec:small}).
\emph{Descriptors} (Fig.~S2). Where structure is used, the entropy coefficient is
negative almost everywhere, so under-confident GNNs sharpen their most uncertain predictions
more. On Computers, relative confidence and ``distance $\geq3$'' carry the largest positive
weights, which softens nodes that are more confident than their neighbours or far from labelled
nodes. Several descriptors are near-duplicates: confidence and entropy, the two LP scores, and neighbour
agreement and similarity correlate at $|r|=0.90$--$0.96$ (variance inflation factors up to 24), so their
weights trade off between splits (33 of 154 setting--descriptor coefficients keep one non-zero sign across
the five splits, entropy most often; Section~S7). The resulting temperatures are the more reliable object to
interpret, and we test them next.

\emph{Faithfulness.} Grouping test nodes by each descriptor, the gap left by one global
temperature (confidence minus accuracy) and \method{}'s mean change of log temperature have
Spearman $\rho=0.67$ over 246 setting--group cells (Fig.~\ref{fig:faith}); where the gap exceeds 0.5 points the temperature
moves in the correcting direction in 81\% of cells, and the mean absolute gap falls from 0.96 to
0.67 points (Section~S11). Nodes three or more hops from labels on Computers and Photo, 16--27 points
over-confident after TS, are softened most, which reduces their gap to 8--18 points.

\subsection{Small Calibration Sets}
\label{sec:small}
On the citation graphs we keep 10\%, 25\% or 50\% of each calibration fold. This leaves about
14/34/67 nodes on Cora, 17/42/83 on Citeseer and 98/246/492 on Pubmed (Fig.~S1).
With so few labels no node-wise calibrator improves on TS (WATS was not run). The NLL of \method{} approaches
that of TS as the set grows (0.827 vs.\ 0.643 at 10\%, 0.679 vs.\ 0.613 at 25\%) and matches it at 50\%
(0.604 vs.\ 0.607; all methods in Section~S1). A practical rule is to use TS or ETS below about 100
calibration nodes.

\begin{table}[t]
\centering
\caption{Per-point calibration of DGCNN on the 2874 ShapeNetPart test shapes (5.8M points), trained without (``original'') or with label smoothing ($\epsilon=0.2$). Calibrators are fitted on 400 validation shapes; mean over 3 backbones $\times$ 5 calibration draws. ECE and AURC in \%, NLL and Brier $\times100$. $^{\downarrow}$/$^{\uparrow}$: \method{} better/worse, 95\% shape-level paired bootstrap interval of the difference excludes zero. RBS follows the original definition; HTS and LTS use the \method{} fitting pipeline. VS changes predictions; all others preserve accuracy.}
\label{tab:pc}
\footnotesize\setlength{\tabcolsep}{1.3pt}
\begin{tabular}{lccccccccc}
\toprule
& \multicolumn{4}{c}{original} & & \multicolumn{4}{c}{label smoothing}\\ \cmidrule(lr){2-5}\cmidrule(lr){7-10}
Method & ECE & NLL & Brier & AURC & & ECE & NLL & Brier & AURC\\
\midrule
Uncal. & 0.64 & 16.89 & 9.13 & 0.773 &  & 1.03 & 18.15 & 9.41 & 0.868\\
TS & 0.26$^{\uparrow}$ & 16.83$^{\downarrow}$ & 9.12$^{\downarrow}$ & 0.773 &  & 0.79$^{\downarrow}$ & 17.86$^{\downarrow}$ & 9.45$^{\downarrow}$ & 0.862\\
VS & 0.45$^{\downarrow}$ & 18.51$^{\downarrow}$ & 9.34$^{\downarrow}$ & 0.885 &  & 0.93$^{\downarrow}$ & 19.07$^{\downarrow}$ & 9.72$^{\downarrow}$ & 1.090\\
ETS & 0.26$^{\uparrow}$ & 16.83$^{\downarrow}$ & 9.12$^{\downarrow}$ & 0.773 &  & 0.79$^{\downarrow}$ & 17.86$^{\downarrow}$ & 9.45$^{\downarrow}$ & 0.862\\
HTS & 0.26$^{\uparrow}$ & 16.83$^{\downarrow}$ & 9.12$^{\downarrow}$ & 0.773 &  & \textbf{0.42}$^{\uparrow}$ & 17.72$^{\downarrow}$ & 9.41$^{\downarrow}$ & 0.856\\
LTS & 0.27$^{\uparrow}$ & 16.82$^{\downarrow}$ & 9.11$^{\downarrow}$ & 0.773 &  & 0.62$^{\downarrow}$ & 17.75$^{\downarrow}$ & 9.42$^{\downarrow}$ & 0.852\\
RBS & \textbf{0.25}$^{\uparrow}$ & 16.81$^{\downarrow}$ & \textbf{9.10}$^{\downarrow}$ & 0.772 &  & 0.58 & 17.74$^{\downarrow}$ & 9.38$^{\downarrow}$ & 0.853\\
CaGCN & 0.61$^{\downarrow}$ & 16.93$^{\downarrow}$ & 9.14$^{\downarrow}$ & 0.775 &  & 1.03$^{\downarrow}$ & 17.91$^{\downarrow}$ & 9.47$^{\downarrow}$ & 0.862\\
\method{} & 0.32 & \textbf{16.80} & \textbf{9.10} & \textbf{0.769} &  & 0.56 & \textbf{17.55} & \textbf{9.33} & \textbf{0.835}\\
\bottomrule
\end{tabular}
\end{table}

\subsection{Application: Point-Cloud Segmentation for Robot Perception}
\label{sec:pc}
A robot that grasps or navigates acts on per-point part predictions and must decide which points
to trust and which to reject or re-observe~\citep{sunderhauf2018limits}. Calibrated confidence
supports this decision, and the area under the risk--coverage curve (AURC)~\citep{geifman2019aurc}
measures it directly: how well confidence ranks points for rejection. We test whether \method{}
transfers, keeping the temperature model and adapting only the descriptors.

\textbf{Setup.} We train DGCNN~\citep{wang2019dgcnn} (without the spatial transformer, 100 epochs,
three seeds) for part segmentation on ShapeNetPart~\citep{yi2016shapenetpart}; test instance mIoU is
84.3\%. Each shape (2048 points) becomes a 10-NN graph. Calibrators are fitted on 400 random
validation shapes (5 draws per seed) or on all 1870, and evaluated on all 2874 test shapes. LP and
distance descriptors do not apply without labels inside a shape, so \method{} uses seven
descriptors (eight parameters): confidence, entropy, neighbour agreement, relative confidence,
neighbour similarity, log mean $k$-NN distance and mean shape confidence. Its cross-validation folds and the
one-standard-error rule of Eq.~(\ref{eq:onese}) work on whole shapes rather than points. GATS needs hop distances
to labelled nodes, and GETS and HoTS were not adapted; RBS uses the soft support ratio on the
$k$-NN graph. We bootstrap test shapes (2000 resamples shared by all methods) and mark a difference
when its 95\% interval excludes zero; the intervals are conditional on the three backbones, so we
also give the sign of each per-seed difference (Section~S5).

\textbf{Results.} The original DGCNN is already well calibrated (ECE 0.64\%, fitted $T\approx1.06$;
Table~\ref{tab:pc}). \method{} gives the best NLL, Brier score and AURC. The NLL and Brier gains are
significant and hold in every seed (except NLL against LTS), and \method{} ranks points for rejection best
in every seed, also when all 1870 validation shapes are used. Its ECE is slightly higher than that of the
single-temperature methods, and the gain is concentrated on interior points (Section~S8).

\textbf{Under-confident backbone.} The official DGCNN code trains with label smoothing; with
$\epsilon=0.2$ the network becomes under-confident. A single temperature then over-corrects: after TS,
confidence lies 0.68 points \emph{above} accuracy, after \method{} 0.45. \method{} beats every baseline on
NLL and Brier score, significantly and in all three seeds, has the lowest AURC in every seed, and its gain
now reaches part boundaries. On ECE it improves on TS and LTS, matches RBS and trails HTS.

\textbf{Sensor degradation.} Calibrators fitted on clean shapes are applied to test shapes with
Gaussian jitter or random point removal (Table~S13). \method{} keeps the lowest NLL, Brier score and
mean AURC under every corruption; most of the gain comes from two of the three backbones (Section~S9).

On the handles and cap peaks a robot would grip, \method{} also has the lowest AURC, with the
gain concentrated on mug handles (Section~S10).

\section{Limitations}
\label{sec:limits}
\method{} needs enough calibration data: below about 100 nodes TS is better, and on ECE GATS remains
ahead. Single coefficients vary across splits when descriptors are collinear. Only one
heterophilous graph was tested, and the hand-crafted descriptors limit the patterns captured. WATS is trained with a stopping rule that differs from its official code (both rules: Section~S12), SimCalib~\citep{tang2024simcalib} and Simi-Mailbox~\citep{seo2025simimailbox} are
missing, and GETS (a port that trains its degree embeddings) was not re-tuned, so a comparison with more
expressive calibrators under a shared protocol is still open. The point-cloud study is a proxy evaluation on CAD
shapes with one backbone and synthetic corruptions; its effect on grasping, navigation and real sensors
remains to be tested.

\section{Conclusion}
\label{sec:conclusion}
\method{} is a lightweight structural calibrator: twelve parameters, fitted in under a second, turn known
causes of GNN miscalibration into a node-wise temperature. The structural descriptors drive its gain, as
the same pipeline fed only the entropy or the logits gains less than half as much NLL over TS. Its best
average rank in NLL and Brier score held on four graphs under a protocol fixed before any run, and on unseen
nodes its temperatures move in the direction of the miscalibration that a global temperature leaves. Future work includes
heterophilous graphs, real sensor data, and combining the shrinkage towards TS with more expressive
calibrators.

\section*{CRediT authorship contribution statement}
\textbf{Chenyi Huang:} Conceptualization, Methodology, Software, Validation, Formal analysis,
Investigation, Data curation, Visualization, Writing -- original draft, Writing -- review \& editing.

\section*{Declaration of competing interest}
The author declares that they have no known competing financial interests or personal relationships
that could have appeared to influence the work reported in this paper.

\section*{Funding}
This research did not receive any specific grant from funding agencies in the public, commercial, or
not-for-profit sectors.

\section*{Data availability}
All datasets are public (Planetoid, Amazon, Coauthor, CitationFull, WikiCS and Amazon-ratings graphs of
PyTorch Geometric, ShapeNetPart). The code, the run-level results and the scripts that generate every table are provided
as supplementary data.

\section*{Declaration of generative AI and AI-assisted technologies in the manuscript preparation process}
During the preparation of this work the author used Claude (Anthropic) and ChatGPT (OpenAI) to
help design and run the experiments and to draft and edit the text.
After using these tools, the author reviewed and edited the content as needed and takes full
responsibility for the content of the published article.

\bibliographystyle{elsarticle-num-names}
\bibliography{refs}

\end{document}